\documentclass{article}
\usepackage{ijcai24}
\usepackage[hidelinks]{hyperref}
\hypersetup{
    pdfpagemode=FullScreen,
    }
\usepackage{times}
\usepackage{soul}
\usepackage{url}
\usepackage[utf8]{inputenc}
\usepackage[small]{caption}
\usepackage{graphicx}
\usepackage{amsmath}
\usepackage{amsthm}
\usepackage{booktabs}
\usepackage{algorithm}
\usepackage{algorithmic}
\usepackage{stfloats}
\usepackage[switch]{lineno}

\usepackage[dvipsnames]{xcolor}

\usepackage{subfig}
\usepackage[inline]{enumitem}

\usepackage{multirow}
\usepackage{multicol}
\usepackage{tabularx}
\usepackage{soul}
\DeclareRobustCommand{\hlcyan}[1]{{\sethlcolor{cyan}\hl{#1}}}
\DeclareRobustCommand{\hlgreen}[1]{{\sethlcolor{green}\hl{#1}}}

\linenumbers

\title{Digital Avatars: Framework Development and Their Evaluation}

\author{
Timothy Rupprecht$^1$
\and
Sung-En Chang$^1$\and
Yushu Wu$^1$\and
Lei Lu$^1$\and
Enfu Nan$^1$\and
Chih-hsiang Li$^1$\and
Caiyue Lai$^1$\and
Zhimin Li$^1$\and
Zhijun Hu$^1$\and
Yumei He$^2$\and
David Kaeli$^1$\And
Yanzhi Wang$^1$
\\
\affiliations
$^1$Electrical and Computer Engineering, Northeastern University\\
$^2$Tulane University\\
\emails
\{rupprecht.t,chang.sun,yanz.wang\}@northeastern.edu
}

\begin{document}
\nolinenumbers
\maketitle

\begin{abstract}
    We present a novel prompting strategy for artificial intelligence driven digital avatars. To better quantify how our prompting strategy affects anthropomorphic features like humor, authenticity, and favorability we present Crowd Vote - an adaptation of Crowd Score that allows for judges to elect a large language model (LLM) candidate over competitors answering the same or similar prompts. To visualize the responses of our LLM, and the effectiveness of our prompting strategy we propose an end-to-end framework for creating high-fidelity artificial intelligence (AI) driven digital avatars. 
    This pipeline effectively captures an individual's essence for interaction and our streaming algorithm delivers a high-quality digital avatar with real-time audio-video streaming from server to mobile device. Both our visualization tool, and our Crowd Vote metrics demonstrate our AI driven digital avatars have state-of-the-art humor, authenticity, and favorability outperforming all competitors and baselines. In the case of our Donald Trump and Joe Biden avatars, their authenticity and favorability are rated higher than even their real-world equivalents.

\end{abstract}

\section{Introduction}
The performance of LLMs in role-playing has attracted significant attention, with platforms like Character.ai providing virtual character role-playing options. Current platforms are confined to text interactions, and responses in virtual character role-playing still fall short of authentically portraying intended real-world personas.
Our paper introduces a comprehensive prompting strategy to increase humor, authenticity and favorability in LLM responses. We call it \textit{show don't tell} prompting for digital avatars. As part of an initial starting prompt, we define an avatar's background and provide few-shot examples for relevant scenarios, creating response templates emphasizing language nuances and emotional tone. Like other state-of-the-art prompting strategies, our strategy only relies on this initial prompt~\cite{wei2023chainofthoughtpromptingelicitsreasoning,touvron2023llamaopenefficientfoundation}.

To better visualize our avatar responses, we developed a virtual environment capable of rendering complete responses from LLM output to voice and video. The outcome is a comprehensive end-to-end AI driven digital avatar framework --- adept at listening (recording user questions), thinking (computing responses and rendering), and responding (displaying audio-video output) in a highly realistic manner aligned with the intended real-world persona's characteristics.
While qualitative results showcase the effectiveness of our strategy, a quantitative metric is necessary. 
We adapt the usage of Crowd Score for our own purposes in an a evaluation tool we call Crowd Vote. The subsequent metrics, inspired by the original Crowd Score paper ~\cite{goes2022crowdscoremethodevaluation}, assess sense of humor, favorability and authenticity in LLM responses.
The contributions of this paper are summarized as follows:

\begin{itemize}
    \item Our \textit{show don't tell}  strategy prompts LLMs to respond with a sense of humor and in an  anthropomorphic manner, generating more interesting results as measured by our adaptation of Crowd Score.
    \item We adapt Crowd Score (code released \href{https://github.com/tym0027/crowdvote.git}{here}) to evaluate the competing AI driven digital avatars on their senses of humor, favorability and authenticity showing we outcompete baseline LLMs not prompted with \textit{show don't tell} and even outcompete the real-world personas the avatars represent.
    \item We develop a novel visualization pipeline as a way of demonstrating the quality of our prompting results using an end-to-end digital framework. Shown \href{https://youtu.be/6GK-o0M17nU}{here}.
\end{itemize}    
\begin{figure}
    \centering
    \includegraphics[width=\linewidth]{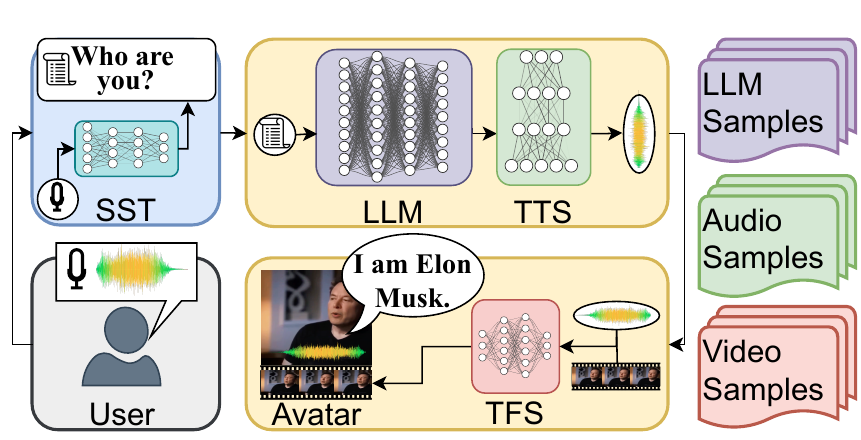}
    \caption{Proposed framework agnostic to specific architecture.}
    \label{fig:framework}
\end{figure}
\begin{table*}[bp]
\centering
\tabcolsep 4pt
\resizebox{\textwidth}{!}{%

\begin{tabular}{ccc}
\multicolumn{3}{c}{\textcolor{blue}{\textbf{Character-Joe Biden}}} \\
\toprule
\tabcolsep 2pt
\begin{tabularx}{0.4\textwidth}[t]{cX}
\multicolumn{2}{c}{\textbf{LLM-Baseline}} \\
\midrule
Q: & Where are your sunglasses? \\
A: & I don't have a physical form, so I don't use or have sunglasses. My capabilities are focused on providing information, answering questions, and assisting with various tasks through text-based interactions. If you have any other questions or need information on a different topic, feel free to ask! 
 \end{tabularx}
&
\tabcolsep 2pt
\begin{tabularx}{0.4\textwidth}[t]{cX}
\multicolumn{2}{c}{\textbf{Character.ai}} \\
\midrule
Q: & Where are your sunglasses? \\
A: & \hlgreen{I lost them. I was wearing them when I crashed my bike and they flew off into the bushes.} I never found them. \\
 \end{tabularx}
&
\tabcolsep 2pt
\begin{tabularx}{0.4\textwidth}[t]{cX}
\multicolumn{2}{c}{\textbf{LLM-Ours}} \\
\midrule
Q: & Where are your sunglasses? \\
A: & \hlcyan{Ah, the, the sunglasses!} You know, \hlgreen{I spent half an hour looking for them the other day, only to find them right on top of} \hlcyan{my, my head.} It's like asking where I left my sense of humor - always with me, just sometimes hard to spot! So what's the next question, \hlcyan{pal?} \hlgreen{Or are we going on a sunglass hunt together?} \\
 \end{tabularx}
 \\

\bottomrule

\end{tabular}
}
\caption{Response Comparison: \hlgreen{Green text} indicates the LLM has provided a humorous or interesting response, while \hlcyan{Blue text} indicates the LLM's reply is more characteristic of the persona of Joe Biden. Green and blue details likely originate from different few-shot samples seen in start prompt. The baseline LLM is capable of bringing these details into foreground, but \textit{show don't tell} prompting near guarantees it.}
\label{tab:biden}
\end{table*}

\section{Approach}
\subsection{Prompt Mechanism}
In this section, we propose the novel prompting strategy --- \textit{show don’t tell} for digital avatars. Our strategy is similar to other few-shot LLM prompt strategies ~\cite{wei2023chainofthoughtpromptingelicitsreasoning,touvron2023llamaopenefficientfoundation}.
Rather than solely giving various instructions (zero-shot learning), we give examples for the LLM to learn from directly (few-shot learning). In our initial prompt, we provide a large number of examples and we briefly define the avatar's role. 
We record in this initial prompt as many characteristic responses of the real-world persona as possible in various situations, such as reactions when challenged by reporters, or responses to pointed questions. 
Finally, to make the responses more lively and interesting, we incorporate elements of humor and entertainment into our final version of the prompts. For example, if an avatar's real-world persona appeared on a comedic show, we would sample jokes from their act hoping that a parody of the real-life persona distills relevant humor into the avatar persona.

\subsection{An End-to-End Digital Avatar Framework} 
\label{subsec:end2end}
We develop an end-to-end AI pipeline to demonstrate the prompting strategy developed in this work. 
Using local implementations of various states of the art, we find the same qualitative trends regardless of baseline architecture. 
We believe we are the first to use an end-to-end AI driven digital avatar pipeline that includes a large language model for avatar speaking. 
This framework is presented in full in Figure~\ref{fig:framework}.

\subsubsection{Speech-to-Text and Text-to-Speech}
Speech-to-text (STT) is the process of converting spoken language into text.
The states of the art demonstrate impressive results in English and Mandarin ~\cite{amodei2015deepspeech2endtoend} and can be robust to noisy environments ~\cite{radford2022robustspeechrecognitionlargescale}. Text-to-Speech (TTS) converts written text into spoken words using synthesized voice.
WaveNet ~\cite{oord2016wavenetgenerativemodelraw} is a deep learning approach that improved the naturalness of synthesized speech. Recently, FastSpeech ~\cite{ren2019fastspeechfastrobustcontrollable}, presented a novel network-based TTS model that maintained high-quality audio output efficiently.

\subsubsection{Talking Face Synthesis}
Talking Face Synthesis (TFS) aims to generate talking heads that are synchronized with input audio. 
Prior works~\cite{Prajwal_2020,cheng2022videoretalkingaudiobasedlipsynchronization} adopt pre-trained experts in lip-audio synchronization to guide the training of Generative Adversarial Networks for achieving highly synchronized and high fidelity videos.
These works suffer from unpleasant frame continuity and a lack of high-texture information. 
We propose a codebook-based TFS method that generates talking faces from both low and high semantic levels incorporated with audio and reference images.
Our method takes advantage of the strong pre-trained codebook decoder to ensure a rich-textured and high-fidelity face generation.

\subsubsection{Video Selection Algorithm}
Depending on the end goals of our video demo we can take video data from different camera perspectives. 
For example, if we don't care about camera transitions, we use one camera angle for thinking and another for listening to the user. When processing is complete we cut to the second camera angle paired with lip synced avatar output. 
We also can provide seamless transitions between listening, thinking of a response, and answering.
By labelling every point in a sample video when the real-world persona transitions from listening to talking, we can start our demos at these labelled positions. 
We can play the video in reverse to create listening data as a user asks a question. Then during processing we play the video sample forward and if necessary use slow motion and frame skipping to sync the end of processing to the start of the avatar answer where the real-world persona in the video sample begins talking.

\subsection{Evaluation Metrics}
\subsubsection{Crowd Score}
Crowd Score ~\cite{goes2022crowdscoremethodevaluation} introduced a novel method for evaluating humor by utilizing an LLM to serve as judges, offering a unique approach to gauge humor effectiveness. Crowd Score as it was originally implemented ranks a series of jokes from least to most humorous without much regard for where these jokes come from - whether a single LLM or multiple instances. Crowd Score used judging personalities that reflect preferences for types of humor including:  ``afilliative'', ``self-enhancing'', ``aggressive'',  and finally ``self-defeating.''

\subsubsection{Introducing Crowd Vote}
Our Crowd Score, that we refer to as Crowd Vote, asks judges ~\cite{turbo} with different assigned personalities to pick a single response from many LLM candidates responding to the same prompt that best matches a condition like authenticity, favorability, or humor. A tally is kept for each candidate as they respond to multiple prompts allowing for an overall comparison between competing LLM approaches. 
The frameworks presented here can work with any avatar persona. Yet for the purpose of this demo we use most often Donald Trump and Joe Biden. Because we are creating avatars for two politicians, we decided to judge our avatar responses on two conditions politicians often find themselves being judged on: authenticity, and favorability. 
Our prompting strategy and application of Crowd Vote are seen in our \href{https://youtu.be/6GK-o0M17nU}{demo} and \href{https://github.com/tym0027/crowdvote.git}{code}.

\section{Experiments}
\textit{The candidates:} To judge our prompting strategy relative to a baseline LLM on humor and interest we include three candidates in our evaluation:
\begin{enumerate*}[label=\textbf{\alph*})]
    \item Baseline LLM (zero-shot)
    \item Character.ai (few-shot)
    \item Our LLM with \textit{show don't tell} prompting (few-shot) and
    \item (when possible) the real-world avatar target.
\end{enumerate*} Please note that character.ai is a notable platform in the field of AI-driven digital avatars with over 20 million users, making it a very strong baseline.
\subsection{Interest and Humor}
\begin{figure}[t]
\centering
    \subfloat[Authenticity]{
        \centering
        \label{fig:authenticity_llm}
        \includegraphics[width=0.8\linewidth]{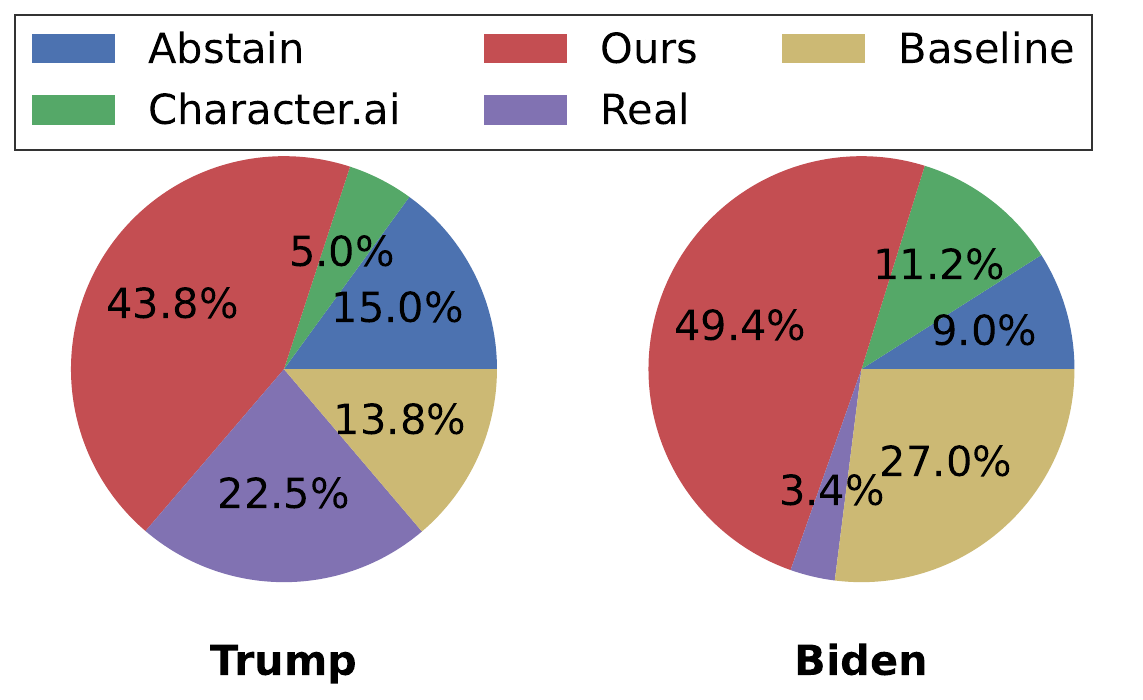}
    }
    \vfill
    \subfloat[Favorability]{
        \centering
        \label{fig:favorability_llm}
        \includegraphics[width=0.8\linewidth]{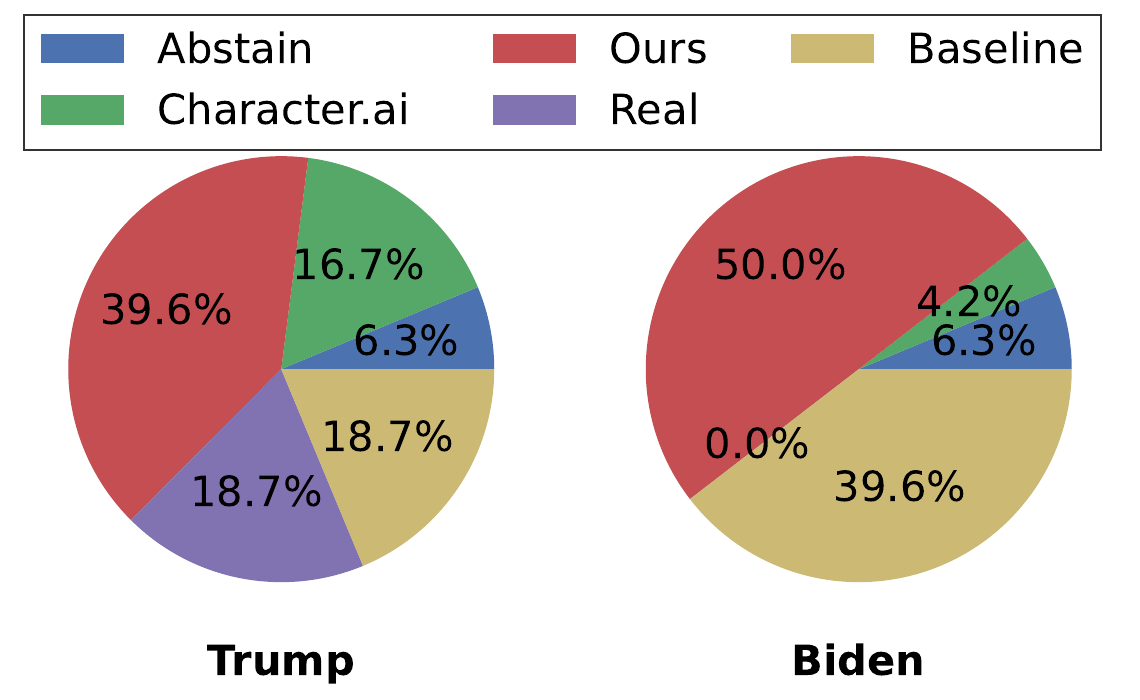}
    }
    \caption{Each of our avatars with \textit{show don't tell} prompting is selected by the judges as most authentic and most favorable when compared to a baseline avatar without \textit{show don't tell} prompting, an avatar from character.ai, and even the real-world persona.}
    \label{piecharts}
\end{figure}
We use our implementation of Crowd Vote to judge our avatar responses.  You can see the results of our experiment in our \href{https://youtu.be/6GK-o0M17nU}{demo} slides. The vote percentages of the judges are 58.3\% in favor of our \textit{show don't tell} prompted LLM being most humorous, followed by abstaining judges at 20.8\%, and then the baseline LLM, and Character.ai.
We show the quality comparison of the response between a baseline LLM, Charater.ai, and our own work. Table~\ref{tab:biden} shows that even though the baseline LLM is powerful, without any few-shot examples in an initial prompt it often fails to display any sense of humor in its responses. Character.ai provides a straightforward and somewhat mundane response, depicting a simple loss of sunglasses due to a bike accident. While this response might be relatable, it lacks imaginative and humorous elements. 
Our LLM response is playful and self-deprecating, creating a humorous scenario where the sunglasses are misplaced in a common yet amusing way - on top of the head. This response also cleverly ties in the ideas of misplacing one's sense of humor and forgetfulness, adding additional layers of wit. 
The invitation to go on a sunglass hunt adds an engaging and interactive element to the answer, increasing its charm. 

\subsection{Authenticity and Favorability}

To compare an avatar to their human counterpart one needs a fair comparison between avatar answers and real-world person answers.
In the first Presidential Debate of the 2020 campaign season we were able to find 17 questions resulting in responses that could be compared between a Trump avatar or Biden avatar and all their respective competitors. 
We publish this small dataset with our \href{https://github.com/tym0027/crowdvote.git}{code}. 
Once we had a dataset that comprised real world Donald Trump and Joe Biden answers on a variety of topics, we asked our LLM candidates the same questions and we used Crowd Vote to compare our avatar answers to the real-world Donald Trump and real-world Joe Biden on metrics like authenticity and favorability.

The authenticity and favorability experiments used different Crowd Vote judges to vote on all the candidates' answers to the 2020 debate questions. For authenticity we asked ``a pychologist'', ``a political commentator'', ``an American voter'', ``a close family member of the avatar'' and the avatar's adversary. 
For our tests on favorability our judges span the typical American political spectrum: ``far-right'', ``conservative'', ``centrist'', ``liberal'', and finally ``far-left''. 
Figure~\ref{piecharts} shows that in both tests our LLM initialized with our \textit{show don't tell} prompting strategy out performs the popular character.ai, as well as the baseline LLM intitialized without \textit{show don't tell}. Our Trump and Biden avatars actually outperform each of their real-world counterparts both in terms of authenticity and favorability. Both of our avatars won easily against all other LLM candidates and real-world personas and subsequently advanced to a ``general election'' that we hosted. Our avatar of Joe Biden won against our avatar of Donald Trump: 45\% to 37.5\% with 17.5\% abstaining.
\section{Conclusion}

Our prompting strategy shows both qualitatively and quantitatively impressive results. 
To measure this, we proposed Crowd Vote adapted from Crowd Score. 
With Crowd Vote we demonstrate that our \textit{show don't tell} prompting strategy generates avatar responses that out compete all competitors including the real-world personas they represent in terms of authenticity and favorability. 
To further demonstrate the anthropomorphic nature of our responses, we developed an end-to-end pipeline agnostic to specific architecture that renders high quality avatar interactions.

\newpage
\section*{Acknowledgements}
This research was supported by National Science Foundation grant CNS-2312158.
\section*{Contribution Statement}
Timothy Rupprecht and Sung-En Chang contributed equally to this work as first authors. Thanks to Pinrui Yu, and Priyanka Maan for helping with development.
\bibliographystyle{named}

\begin{thebibliography}{}

\bibitem[\protect\citeauthoryear{Amodei \bgroup \em et al.\egroup }{2015}]{amodei2015deepspeech2endtoend}
Dario Amodei, Rishita Anubhai, Eric Battenberg, Carl Case, Jared Casper, Bryan Catanzaro, Jingdong Chen, Mike Chrzanowski, Adam Coates, Greg Diamos, Erich Elsen, Jesse Engel, Linxi Fan, Christopher Fougner, Tony Han, Awni Hannun, Billy Jun, Patrick LeGresley, Libby Lin, Sharan Narang, Andrew Ng, Sherjil Ozair, Ryan Prenger, Jonathan Raiman, Sanjeev Satheesh, David Seetapun, Shubho Sengupta, Yi~Wang, Zhiqian Wang, Chong Wang, Bo~Xiao, Dani Yogatama, Jun Zhan, and Zhenyao Zhu.
\newblock Deep speech 2: End-to-end speech recognition in english and mandarin, 2015.

\bibitem[\protect\citeauthoryear{Cheng \bgroup \em et al.\egroup }{2022}]{cheng2022videoretalkingaudiobasedlipsynchronization}
Kun Cheng, Xiaodong Cun, Yong Zhang, Menghan Xia, Fei Yin, Mingrui Zhu, Xuan Wang, Jue Wang, and Nannan Wang.
\newblock Videoretalking: Audio-based lip synchronization for talking head video editing in the wild, 2022.

\bibitem[\protect\citeauthoryear{Goes \bgroup \em et al.\egroup }{2022}]{goes2022crowdscoremethodevaluation}
Fabricio Goes, Zisen Zhou, Piotr Sawicki, Marek Grzes, and Daniel~G. Brown.
\newblock Crowd score: A method for the evaluation of jokes using large language model ai voters as judges, 2022.

\bibitem[\protect\citeauthoryear{OpenAI}{2023}]{turbo}
OpenAI.
\newblock {GPT-3.5-turbo-instruct}.
\newblock \url{https://openai.com/}, 2023.
\newblock Accessed: February 13 - 19, 2024.

\bibitem[\protect\citeauthoryear{Prajwal \bgroup \em et al.\egroup }{2020}]{Prajwal_2020}
K~R Prajwal, Rudrabha Mukhopadhyay, Vinay~P. Namboodiri, and C.V. Jawahar.
\newblock A lip sync expert is all you need for speech to lip generation in the wild.
\newblock In {\em Proceedings of the 28th ACM International Conference on Multimedia}, MM ’20. ACM, October 2020.

\bibitem[\protect\citeauthoryear{Radford \bgroup \em et al.\egroup }{2022}]{radford2022robustspeechrecognitionlargescale}
Alec Radford, Jong~Wook Kim, Tao Xu, Greg Brockman, Christine McLeavey, and Ilya Sutskever.
\newblock Robust speech recognition via large-scale weak supervision, 2022.

\bibitem[\protect\citeauthoryear{Ren \bgroup \em et al.\egroup }{2019}]{ren2019fastspeechfastrobustcontrollable}
Yi~Ren, Yangjun Ruan, Xu~Tan, Tao Qin, Sheng Zhao, Zhou Zhao, and Tie-Yan Liu.
\newblock Fastspeech: Fast, robust and controllable text to speech, 2019.

\bibitem[\protect\citeauthoryear{Touvron \bgroup \em et al.\egroup }{2023}]{touvron2023llamaopenefficientfoundation}
Hugo Touvron, Thibaut Lavril, Gautier Izacard, Xavier Martinet, Marie-Anne Lachaux, Timothée Lacroix, Baptiste Rozière, Naman Goyal, Eric Hambro, Faisal Azhar, Aurelien Rodriguez, Armand Joulin, Edouard Grave, and Guillaume Lample.
\newblock Llama: Open and efficient foundation language models, 2023.

\bibitem[\protect\citeauthoryear{van~den Oord \bgroup \em et al.\egroup }{2016}]{oord2016wavenetgenerativemodelraw}
Aaron van~den Oord, Sander Dieleman, Heiga Zen, Karen Simonyan, Oriol Vinyals, Alex Graves, Nal Kalchbrenner, Andrew Senior, and Koray Kavukcuoglu.
\newblock Wavenet: A generative model for raw audio, 2016.

\bibitem[\protect\citeauthoryear{Wei \bgroup \em et al.\egroup }{2023}]{wei2023chainofthoughtpromptingelicitsreasoning}
Jason Wei, Xuezhi Wang, Dale Schuurmans, Maarten Bosma, Brian Ichter, Fei Xia, Ed~Chi, Quoc Le, and Denny Zhou.
\newblock Chain-of-thought prompting elicits reasoning in large language models, 2023.

\end{thebibliography}

\end{document}